\documentclass[letterpaper, 10 pt, conference]{ieeeconf}
\IEEEoverridecommandlockouts
\usepackage{listings}
\usepackage{xcolor}
\usepackage{graphicx}
\usepackage{tcolorbox}
\usepackage{mdframed}
\usepackage{amsmath, amssymb}
\usepackage{newtxtext,newtxmath}
\usepackage{etoolbox}
\usepackage{caption} 
\usepackage{cite}
\title{Grounded Task Axes: Zero-Shot Semantic Skill Generalization via Task-Axis Controllers and Visual Foundation Models}

\author{M. Yunus Seker$^{1}$, Shobhit Aggarwal$^{1}$, and Oliver Kroemer$^{1}$
\thanks{$^{1}$Carnegie Mellon University, The Robotics Institute,
        Pittsburgh, PA, USA
        {\tt\small mseker@andrew.cmu.edu}}%
}

\begin{document}

\makeatletter
\def\@maketitle{%
  \newpage%
  \null%
  \begin{center}%
    {\LARGE \@title \par}%
    \vskip 1em%
    {\large \@author \par}%
    \vskip 1em%
    \includegraphics[width=\textwidth]{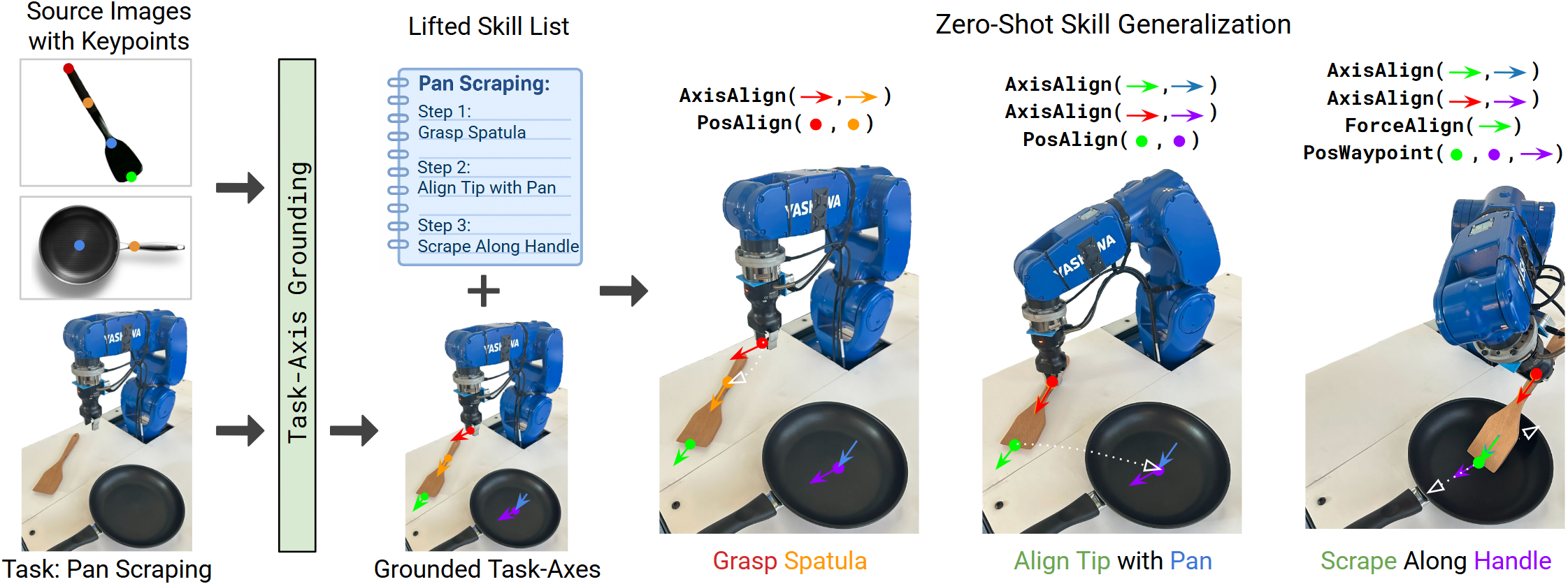}
    \captionof{figure}{Grounded Task-Axes (GTAs): Our system generalizes task-axis controllers zero-shot via vision-based keypoint correspondences.}
    \label{fig:teaser}
    \vspace{-0.5cm}
  \end{center}%
}
\makeatother

\maketitle
\thispagestyle{empty}
\pagestyle{empty}
\begin{abstract}
Transferring skills between different objects remains one of the core challenges of open-world robot manipulation. Generalization needs to take into account the high-level structural differences between distinct objects while still maintaining similar low-level interaction control. 
In this paper, we propose an example-based zero-shot approach to skill transfer. Rather than treating skills as atomic, we decompose skills into a prioritized list of grounded task-axis (GTA) controllers. Each GTAC defines an adaptable controller, such as a position or force controller, along an axis. Importantly, the GTACs are \emph{grounded} in object key points and axes, e.g., the relative position of a screw head or the axis of its shaft. Zero-shot transfer is thus achieved by finding semantically-similar grounding features on novel target objects. We achieve this example-based grounding of the skills through the use of foundation models, such as SD-DINO, that can detect semantically similar keypoints of objects.  We evaluate our framework on real-robot experiments, including screwing, pouring, and spatula scraping tasks, and demonstrate robust and versatile controller transfer for each.
\end{abstract}

\section{Introduction}

Transferring robot manipulation skills across diverse objects remains a fundamental challenge in open-world robotics. Humans intuitively generalize tasks across varied objects by identifying shared semantic and functional characteristics, even when confronted with substantial differences in appearance or geometry. Robotic systems, however, typically lack this intuitive adaptability, often failing to generalize when faced with even minor variations. Addressing this limitation requires robotic systems capable of abstracting tasks at a semantic level while still reliably performing low-level physical interactions.

A promising strategy to address this issue involves decomposing complex manipulation skills into modular components defined relative to object-centric coordinate frames or task-specific axes. Early work introduced structured representations of robot manipulation tasks based on meaningful task frames and axes~\cite{mason1981compliance, ballard1984task, raibert1981hybrid}. Building upon these concepts, Sharma et al.~\cite{sharma2020learning, sharma2021generalizing} further advanced this approach by defining tasks as prioritized sequences of modular controllers. These modular frameworks significantly improved robustness, interpretability, and flexibility in robot manipulation. However, despite their effectiveness, existing task-axis methods still generally require extensive task-specific training, heavily annotated datasets or multiple demonstrations, and predefined sets of controllers, thus limiting their ability to generalize skills in zero-shot.

To overcome these limitations, we propose a novel zero-shot approach to semantic grounding of task-axis controllers. Our framework significantly expands the generalization capabilities of task-axis controllers by leveraging recent advancements in visual foundation models. Specifically, we propose a divide-and-conquer methodology that decomposes complex manipulation tasks into individual grounded task-axis (GTA) controllers. Each GTA controller specifies the low-level robot action—such as force application or positional adjustment—along clearly defined axes and keypoints grounded to the objects' local geometry. Rather than relying on extensive retraining or demonstrations, we enable zero-shot generalization through example-based semantic grounding using powerful visual foundation models like SD-DINO~\cite{sddino}.

Visual foundation models, such as SD-DINO, integrate semantic and geometric information, learned from large-scale pretraining, allowing for robust identification of semantically similar keypoints across highly varied object instances. In our framework, we exploit these capabilities to perform keypoint matching between reference and novel objects purely from RGB-D images. This matching facilitates accurate grounding of task-axis controllers in novel scenarios without task-specific demonstrations or object-specific training. Such a semantic and visual approach contrasts sharply with traditional geometric-based methods, offering enhanced robustness to significant variations in object shapes, textures, colors, and viewing angles.
Our method makes four key contributions to skill generalization in robotic manipulation:

\noindent \textbf{Modular Skill Representation:} We represent complex skills as prioritized sets of grounded task-axis controllers, improving interpretability,providing task-tailored manipulation control, and enabling sub-skill controller reuse.

\noindent \textbf{Semantic Grounding with Visual Foundation Models:} We use SD-DINO to match keypoints across object instances, allowing accurate grounding in novel scenarios.

\noindent \textbf{Zero-Shot Generalization:} Our framework generalizes skills to novel objects without any training or demonstrations.

\noindent \textbf{Real-World Evaluation:} We validate our approach on diverse tasks, pan scraping, pouring, and screwing, demonstrating zero-shot execution across varied object geometries.

\setcounter{figure}{1}
\begin{figure*}[t]
    \centering
    \includegraphics[width=\linewidth]{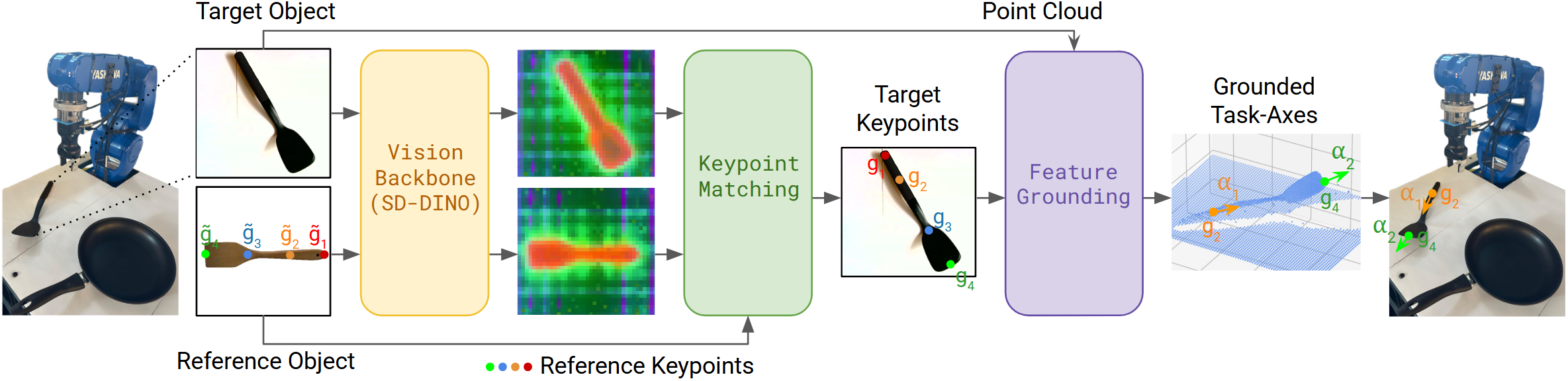}
    \caption{Vision pipeline for grounding task-axes: Given a reference object annotated with keypoints and a target object, GTA uses SD-DINO as a vision backbone to extract visual features. We perform keypoint matching to find corresponding keypoints on the target object. Using these keypoints and 3D point cloud data, we ground the object-centered task-axis controllers in real-world.}
    \label{fig:method}
    \vspace{-0.5cm}
\end{figure*}

\section{Related Work}
\label{sec:related_work}
\textit{Task-Axis Controllers:}
Task-axis frames have long been used in robotic manipulation to define motions relative to meaningful reference frames. Early works~\cite{mason1981compliance, ballard1984task, raibert1981hybrid} introduced constraint-based formulations for specifying manipulation along task-axes. Later methods further developed this idea, defining task-specific motions using fixed task-frames or axes for reliable execution~\cite{manschitz2020learning, migimatsu2020object, king2016rearrangement, berenson2011task}.

More recently, task-axis controllers have been used as modular components for structuring complex behaviors. Sharma et al.\cite{sharma2020learning} proposed composing such controllers via reinforcement learning, and later extended the framework to infer controller parameters from visual input via dense correspondence learning\cite{sharma2021generalizing}. However, these methods depend on task-specific training. In contrast, our method performs inference in a zero-shot manner without retraining.

\textit{Imitation Learning and Trajectory Generation:}
Skill learning  often involves training trajectory generators using imitation learning. Methods like DMPs~\cite{paraschos2013probabilistic, seker2019conditional, ijspeert2013dynamical} often require endpoints to be specified and a suitable coordinate frame to be selected for generalization. Rather than identifying a task-oriented coordinate frame, recent diffusion-based policies~\cite{chi2023diffusion, shridhar2023perceiver}, generate trajectories in global frames. Perceiver-Actor~\cite{shridhar2023perceiver} directly predicts a target gripper frame and uses a motion planner for execution. By mapping scenes to compact keypoints and axes, our method can reduce input dimensionality and improve data efficiency. In the future, we plan to incorporate imitation learning approaches for trajectory generation into our controller library.

\textit{Visual Foundation Models for Correspondences:}
Vision foundation models have shown strong performance in establishing semantic and dense correspondences. Models like CLIP~\cite{clip}, DiFT~\cite{tang2023emergent}, and DINO-ViT~\cite{amir2021deep} learn semantic features from large-scale pretraining and enable pixel-level matching without supervision~\cite{zhu2024densematcher}. We leverage SD-DINO~\cite{sddino} to compute sparse semantic correspondences for robust keypoint matching. Our system remains modular and can incorporate future vision models seamlessly.


\textit{Keypoint Correspondences for Manipulation:}
Recent works like DINOBot~\cite{di2024dinobot}, SKIL~\cite{wang2025skil}, and MAGIC~\cite{liu2024one} extract keypoints from pretrained vision models to enable generalization across object categories. However, these methods train task-specific policies from demonstrations or rely on real-to-sim transfer for planning~\cite{fang2024keypoint, ju2024robo, zhang2023universal}. By contrast, our method operates entirely in a zero-shot setting, without demonstrations or policy training.


\section{Method}
\label{sec:method}

\subsection{Task-Axis Controllers}

Our proposed modular framework creates skills by combining multiple grounded task-axis controllers in parallel. This skill structure builds on the  work of Sharma and Liang et al~\cite{sharma2020learning, sharma2021generalizing}, and we provide a more general formulation for these types of skills. It should be noted that our skills also build upon more classical task-frame formulations of robot manipulation~\cite{mason1981compliance, ballard1984task, raibert1981hybrid}, and our skill formulation is intended to be flexible enough to capture a wide range of skills. 

In this section, we define the notation and structure of individual task-axis controllers. The following sections describe how they are combined to form skills (Sec \ref{subsec:skills}), how we ground them in the objects of new scenes (Sec \ref{subsec:grounding} and \ref{subsec:featuremaps}), and examples of specific controllers used in our experiments (Sec \ref{subsec:controllers}).

A robot skill will consist of $D$ grounded task-axis (GTA) controllers running in parallel. The $i$-th GTA controller has a set of $m_i\geq 0$ keypoints $\mathbf{g}_i=\{g_{i1},...,g_{im}\}$ where $g_{ij}\in\mathbb{R}^3$, and a set of $n_i > 0$ axes $\boldsymbol{\alpha}_i=\{\alpha_{i1},...,\alpha_{in}\}$ where $\alpha_{ij}\in\mathbb{R}^3$. We refer to these keypoints and axes as the \emph{grounding parameters} of the controller, as they form the basis for specifying the controller within the context of the current scene. The grounding process will anchor these features to the geometry of the objects in the scene, as well as the robot (e.g., a keypoint may be associated to the tip of a screw or a point on the gripper).   

Each controller must specify at least one primary axis $\alpha_{i1}$ as the direction in which the controller is acting. For example, the primary axis $a_{i1}$ will define the 3D world frame direction in which a force controller will apply a force. 

The grounding parameters may vary over time, e.g., if the parameter reflects an object that is moving. We denote the time dependency with an additional subscript $t$, i.e., $\mathbf{g}_{it}$=$\{g_{i1t},...,g_{imt}\}$. The variable $\alpha_{i10}$ thus specifies the controller's primary axis as defined at time step zero.

The action selection along the task axis $\alpha_{i0}$ is defined by a controller $\pi_i$ and, sometimes, a trajectory generator $\tau_i$. The trajectory generator is defined as $x^d_t$=$\tau_i(o,\mathbf{g}_{i},\boldsymbol{\alpha}_{i},t;c^\tau_i,\theta^\tau_i)$, where $o$ are sensor observations, $c^\tau$ are fixed parameters that specify the type of trajectory generator, and $\theta^\tau$ specifies a set of tunable parameters for this type of generator. The output $x^d$ specifies a desired value, which is passed to the controller. 

We similarly define the controller policy as  $u_t=\pi_i(o,\mathbf{g}_{i},\boldsymbol{\alpha}_{i},x^d_t;c^\pi_i,\theta^\pi_i)$. To ease notation, we will often merge the trajectory generator and controller jointly as a time-dependent controller $u_t=\pi_i(o,\mathbf{g}_{i},\boldsymbol{\alpha}_{i},t;c_i,\theta_i)$ 
In practice,  the controller may often be specified based on the initial set of grounding parameters $u_t=\pi_i(o,\mathbf{g}_{i0},\boldsymbol{\alpha}_{i0},t;c_i,\theta_i)$ or the most recent values $u_t=\pi_i(o,\mathbf{g}_{it},\boldsymbol{\alpha}_{it},t;c_i,\theta_i)$.

\begin{figure*}[t]
    \centering
    \includegraphics[width=\linewidth]{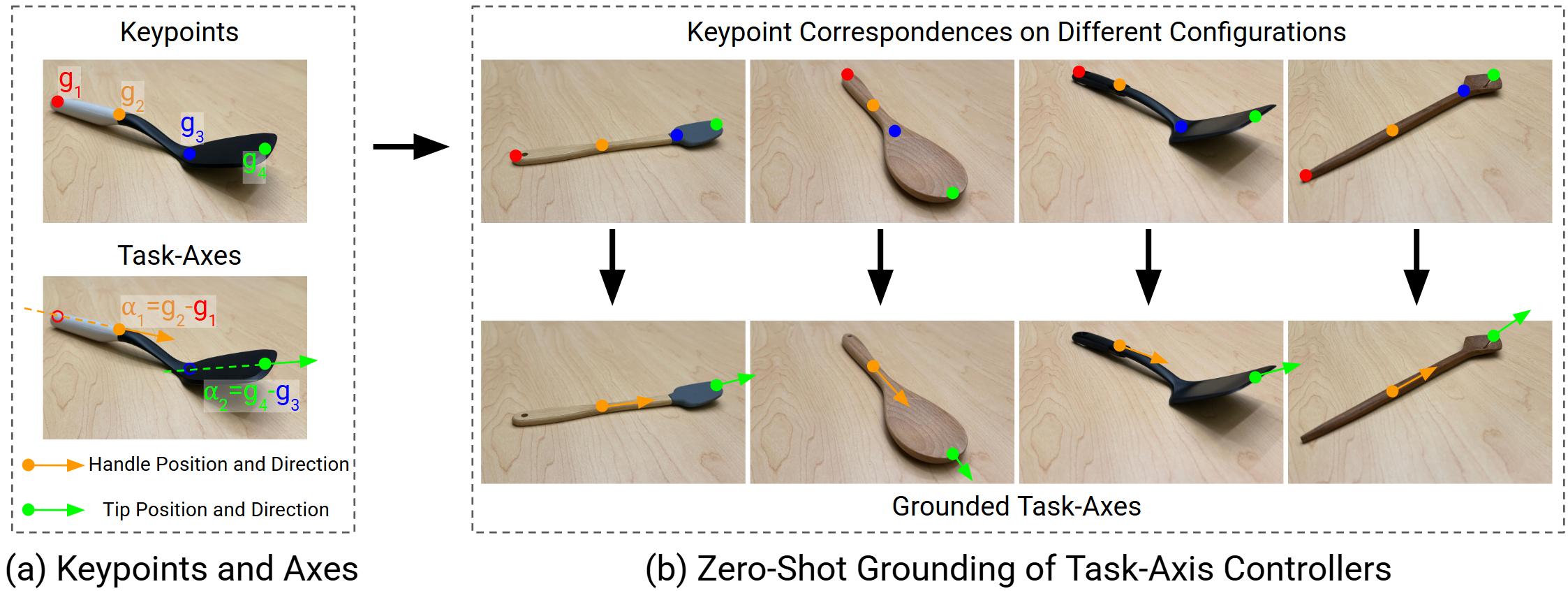}
    \caption{(a) Example Keypoints and Task-Axes for a spatula object. (b) Zero-shot generalization to any object configuration.}
    \label{fig:controller_design_generalization}
    \vspace{-0.5cm}
\end{figure*}

\subsection{Task-Axis Controller Skills}
\label{subsec:skills}

A skill $\pi$ is defined as a prioritized list of GTA controllers that are executed simultaneously. The index of the controller $i$ indicates the ordering in the list, with lower indices indicating higher priority ($i$=$1$ is first priority, $i$=$2$ is second priority, etc.).  

The robot needs to ensure that the controllers do not interfere with the execution of higher-priority controllers. Lower priority controllers are therefore projected into the null space of higher, resulting in a projected normalized axis $\hat\alpha$ of

$$\hat{\alpha}_{i1}=\text{Proj}(\alpha_{i1},\{\alpha_{j1}|\forall j<i\}), ||\hat\alpha_{i1}||=1$$
with corresponding projected action commands $\hat{u}$.  In practice, we use a separate list for translational and rotational controllers.  

Three controller axes specify a full Cartesian coordinate frame. However, lower dimensional skills can also be defined using the framework. For example, to place a screw into a hole, the robot only needs to specify an axis $\alpha_{110}$ along the fixed central axis of the hole and a second $\alpha_{21t}$ radially from the hole to the screw tip. The skill formulation thus reflects the rotational invariant nature of the task. Similarly, for the orientation of the screw, we only need to specify a single control axis $\alpha_{21t}$ for aligning the hole and screw axes. 

The resulting skills are defined in the Cartesian space, as is common for manipulation tasks. Ultimately, any robot skill needs to define a set of low-level joint commands. For completeness, we define this mapping as the \emph{robot implementation} $R$ of the GTA skill. The implementation function $R(\hat{\boldsymbol{\alpha}},\hat{\boldsymbol{u}}, \boldsymbol{c})$ takes in  the list of projected axes $\hat{\boldsymbol{\alpha}}$, the corresponding list of projected actions $\hat{\boldsymbol{u}}$, and the types of controllers $\boldsymbol{c}$ to compute the low-level joint commands. In our experiment, we only used a single type of robot. However, in the future GTA skills could be ported between robots by creating individual implementations for a shared library.

\subsection{Grounding Task-Axis Controllers with Example Keypoints}
\label{subsec:grounding}

The skill formulation in the previous sections allows us to define GTA controllers and skills in an abstract form with unspecified variables for the keypoints and axes. Adopting nomenclature from symbolic planning, we refer to skills for which all of the other parameters ($c$,$\theta$, controller order, etc.) have been specified but the grounding parameters $g$ and $\alpha$ are unspecified as \emph{lifted} task-axis skills. These skill definitions can thus specify manipulations, such as peg insertion, in the abstract. However, to execute the skill for a specific scenario and set of objects, we need to define the parameters through a grounding process. The resulting fully-defined skills then become \emph{grounded} task-axis skills.

Determining the grounding parameters is a critical step in transferring a controller or skill to a new scenario. The grounding process \emph{anchors} the parameters to objects in the scene. 
In this paper, we focus on example-based zero-shot grounding transfer based on RGB-D images $\mathcal{I}$. This approach allows us to leverage existing foundation models to provide broad transfer of skills between semantically similar objects. 

We assume that the robot is provided with example grounding points $\tilde{g}_{ij}$ for a previous source image $\mathcal{I}^\mathcal{S}$. The goal is then to find the corresponding point ${g}_{ij}$ for a new target image $\mathcal{I}$. To ease notation, we assume the camera is calibrated and that the depth value of each pixel in the image $\mathcal{I}$ specifies a full 3D position in the world frame that the robot can index. 

To transfer an example keypoint $\tilde{g}$ and find a similar keypoint for grounding in the new scene, the robot uses feature maps $\phi$ computed by foundation model such as SD-DINO~\cite{sddino}.  The feature map $\phi(\tilde{g}_{ij}, \mathcal{I}_\mathcal{S})\in \mathbb{R}^k$ maps the example point into a $k$ dimensional embedding space. To find the corresponding grounding point in the current target scene $\mathcal{I}$, we define a mapping function $M$ that finds the closest point in the embedding space
$$ g_{ij}= M(\tilde{g}_{ij}, \mathcal{I}^\mathcal{S},\mathcal{I})= \arg\max_{g} \phi(\tilde{g}_{ij}, \mathcal{I}^\mathcal{S})^T \phi(g, \mathcal{I})$$
Note that the range of candidate values for $g$ to maximize over is limited to the 3D points corresponding to the pixels image $\mathcal{I}$. 
The $\arg\max$ operator can be susceptible to noise depending on the quality of the image. We therefore also implemented a soft-max mapping 
$$M(\tilde{g}_{ij}, \mathcal{I}^\mathcal{S},\mathcal{I})= \mathcal{Z}^{-1}\sum_g  \exp(\epsilon^{-1} \phi(\tilde{g}_{ij}, \mathcal{I}^\mathcal{S})^T \phi(g, \mathcal{I}))$$ 
where $\mathcal{Z}$ is a normalization term for the softmax and $\epsilon$ is the temperature term. 

To transfer an example axis $\tilde{\alpha}$ to a new scene, we consider a couple of different mappings depending on the type of axis. For global axes, such as the direction of gravity, we may use an identity mapping $\alpha_{ij}$=$\tilde{\alpha}_{ij}$. For axes defined by keypoints, we can directly use the transferred points to compute the axis $\alpha_{ij}$=$f_{ij}(\boldsymbol{g}_i)$, thus leveraging the keypoint transfer mechanism. Some axes are defined based on local object geometry, e.g., a surface normal or the direction of an edge. To capture these axes, we still require a reference keypoint $\tilde{g}^\alpha_{ik}$ to define the local region from which to extract the axis
$$\alpha_{ij}=f_{ij}(M(\tilde{g}^\alpha_{ij}, \mathcal{I}^\mathcal{S},\mathcal{I}),\mathcal{I})$$
where the mapping function $M$ finds a suitable semantically similar point in the new image, and the function $f$ extracts the axis based on the local geometry in the image $I$ around this mapped point. 
Other forms of axis mappings could be specified in the future. For our experiments, and a wide range of manipulation tasks, the above types of axis mappings are sufficient for creating versatile skills. 

\begin{figure}[b]
    \vspace{-0.5cm}
    \centering
    \includegraphics[width=\linewidth]{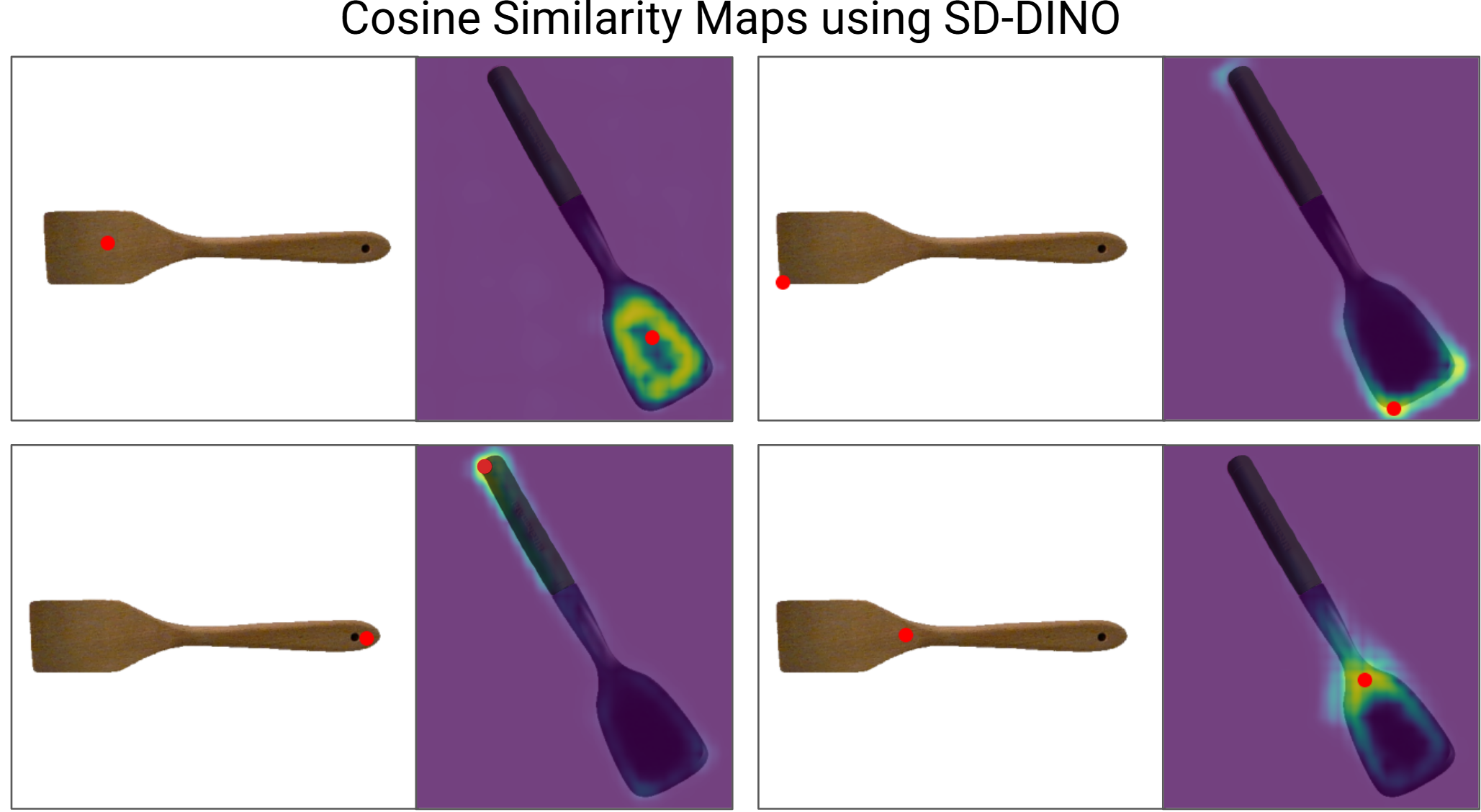}
    \caption{Visualization of cosine similarity maps on target object image according to different pixel keypoints selected on the reference object.}
    \label{fig:sd_dino_matching}
    \vspace{-0.5cm}
\end{figure}

\begin{figure*}[t]
    \centering
    \includegraphics[width=\linewidth]{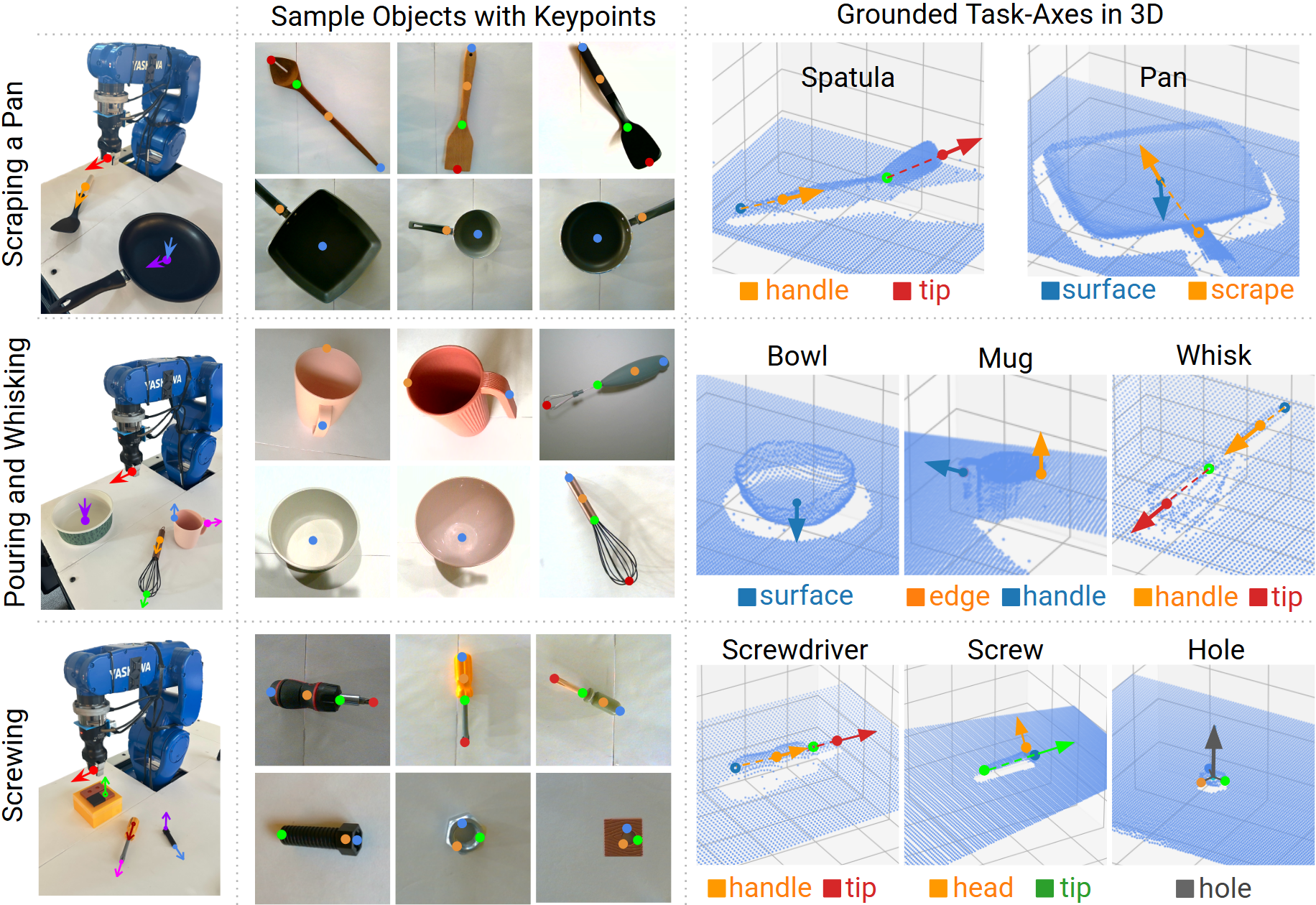}
    \caption{(Left) Target tasks: Scraping a pan, pouring, and screwing. (Middle) Example object configurations with keypoint annotations from our in-lab data collection (Right) Grounded Task-Axes derived from keypoints.}
    \label{fig:experiment_setup}
    \vspace{-0.5cm}
\end{figure*}

\begin{figure*}[t]
    \centering
    \includegraphics[width=\linewidth]{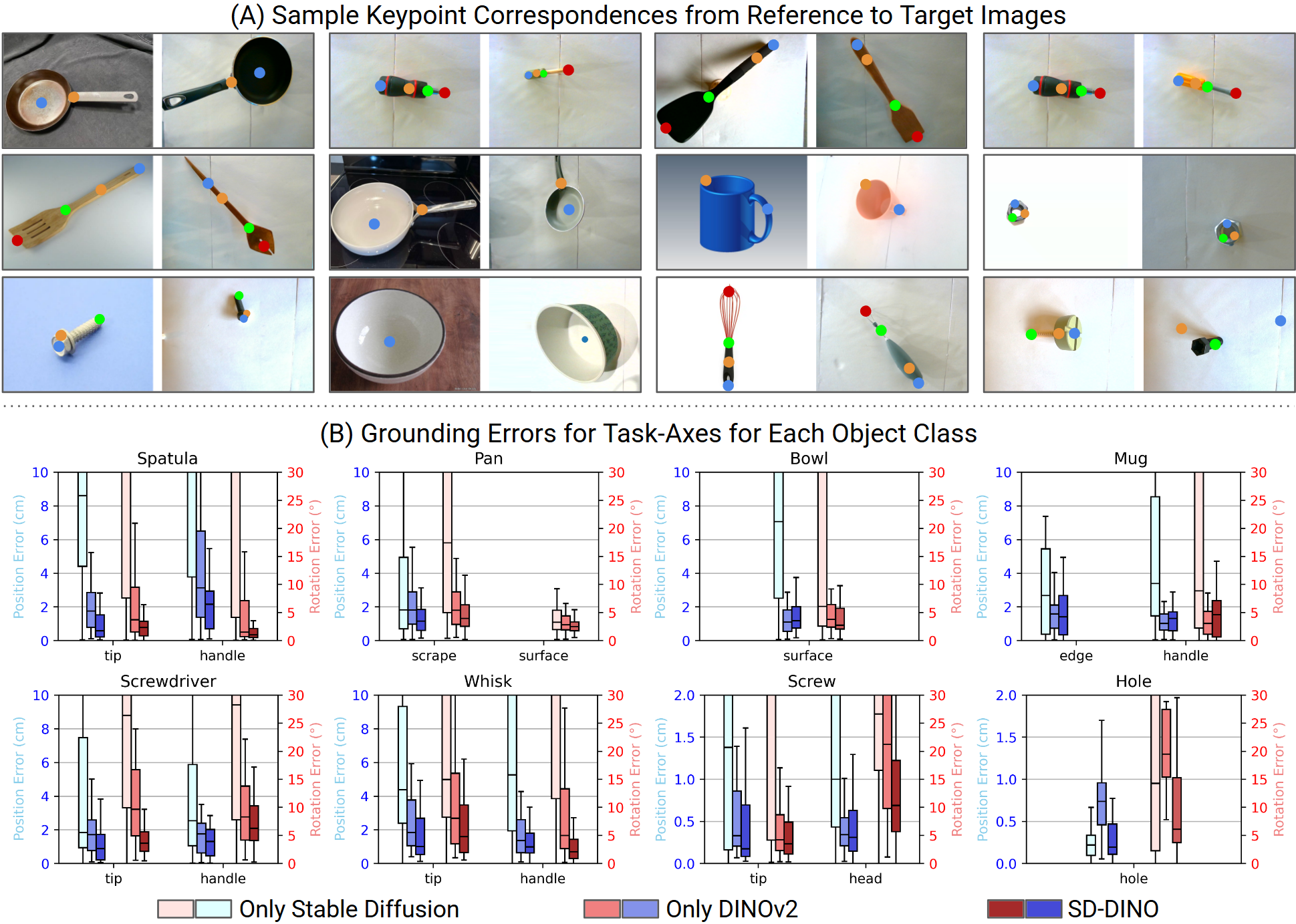}
    \caption{(A) Examples for keypoint correspondences. Left side of the pairs show the reference keypoints, right side of the pairs show the corresponding keypoints using SD-DINO. Failure case in right-bottom. (B) Errors between ground-truth and grounded task-axes for each object class.}
    \label{fig:controller_errors}
    \vspace{-0.5cm}
\end{figure*}

\subsection{Feature Maps}
\label{subsec:featuremaps}

The feature map $\phi$ provides the basis of grounding lifted skills in new scenarios by finding similar points in the new scene to those in the reference example. The feature maps thus define the notion of similarity between points. 

In manipulation tasks, we often need to consider the semantic similarity of keypoints for transfer. For example, tools often have tool tips, like the head of a screwdriver or the tip of a pen, that are critical to their successful usage. We therefore want a feature map that can capture such semantically similar points, despite small local geometric variations (e.g., a screwdriver with a different tip shape) or even larger global geometric variations (e.g., a hand screwdriver versus an L-shaped electrical one). We use the foundation model DINOv2~\cite{oquab2023dinov2} as one of our feature maps, as it is well suited for capturing semantically similar points. 

In some tasks, we also need to consider the spatial location of the points. For example, all edges of a plate may be semantically similar, but we want the robot to grasp the left side of a plate. We therefore want a feature map that can also capture more general spatial information. We achieve this using the Stable Diffusion (SD)~\cite{stablediffusion} network.

To obtain the potential benefits of both approaches, we leverage a visual foundation model SD-DINO~\cite{sddino}, which is a combination of both  DINOv2~\cite{oquab2023dinov2} and Stable-Diffusion~\cite{stablediffusion}. 
We followed the procedure given in the SD-DINO paper to extract pixel-wise features for the given reference and target images. We first calculated the DINOv2 patch token features using DINOv2-b14, and we combined these features with the 3rd, 5th, and 11th decoder layer features of the Stable Diffusion at timestep 50. We upscaled all the features from both models to align their shapes with the input image dimensions. 

To find the corresponding keypoints on the target image for each reference keypoint, we first took an average around each reference keypoint feature using a 3x3 window, and calculated a cosine similarity map using the features of each pixel in the target image. Finally, we selected the best corresponding keypoint by applying a soft-argmax to this cosine map with a temperature of 0.01. Figure~\ref{fig:sd_dino_matching} visualizes an example for the correspondence matching using the SD-DINO cosine similarity maps.

\subsection{Example Controllers}
\label{subsec:controllers}

In our experiments, we use a set of only 4 types of GTA controllers to create the various skills. This set can be expanded in the future. We describe our current library of GTA controllers below:

\medskip\noindent\textbf{$\textsc{PosAlign}(g_{i1},g_{i2},\theta_i)$ -}  defines a position controller that operates along $\alpha_{i1}=g_{i1}-g_{i2}$ to move keypoint $g_{i1}$ on the robot's end-effector or tool to keypoint $g_{i2}$. The controller parameter $\theta_i\in \mathbb{R}^3$ defines a target offset between the two keypoints, such that $\theta_i$=$0$ moves the two points into alignment.  

\medskip\noindent\textbf{$\textsc{PosWaypoint}(g_{i1},g_{i2},\alpha_{i2},\theta_i)$ -}  defines a position controller that moves keypoint $g_{i1}$ on the robot's end-effector or tool to a set of waypoints along $g_{i2}+\alpha_{i1}$. $\theta_i$ specifies a list of offset waypoints.

\medskip\noindent\textbf{$\textsc{AxisAlign}(\alpha_{i1},\alpha_{i2},\theta_i)$ -}  defines a rotational controller that rotates axis $\alpha_{i1}$ on the robot's end-effector or tool to align with axis $\alpha_{i2}$. Similar to PosAlign, $\theta_i\in \mathbb{R}^3$ defines a target offset angle between the axes. 

\medskip\noindent\textbf{$\textsc{ForceAlign}(\alpha_{i1},\theta_i)$ -}  defines a force controller that operates along $\alpha_{i1}$ to align with force value $\theta_i$.

\begin{figure}[b]
    \vspace{-0.5cm}
    \centering
    \includegraphics[width=\linewidth]{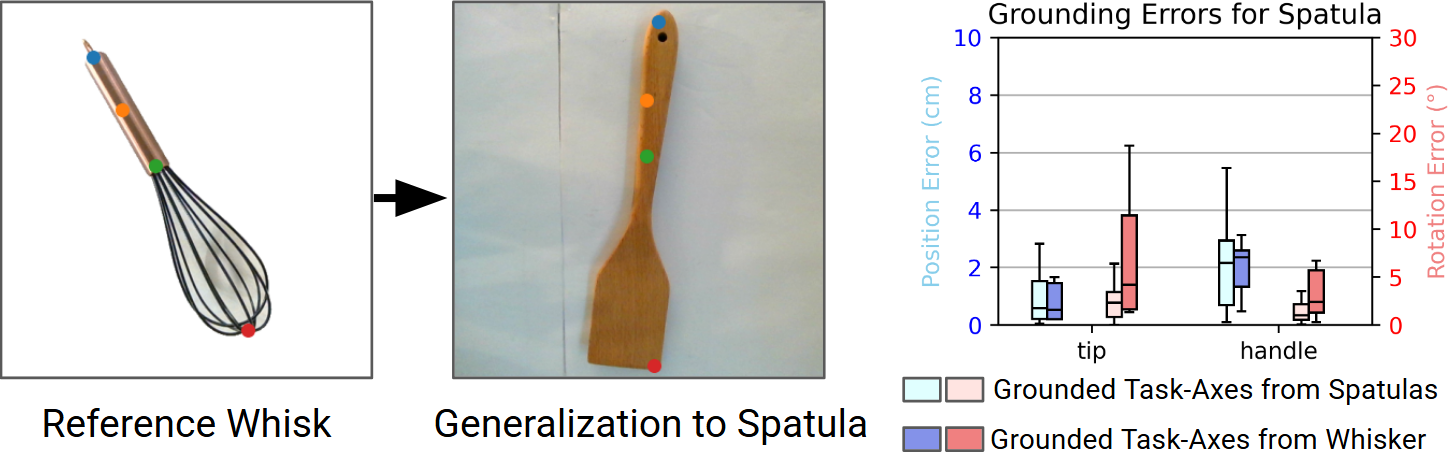}
    \caption{Grounding task-axes from a reference whisk to spatulas.}
    \label{fig:transferring_controllers}
    \vspace{-0.5cm}
\end{figure}

\begin{figure*}[t]
    \centering
    \includegraphics[width=\linewidth]{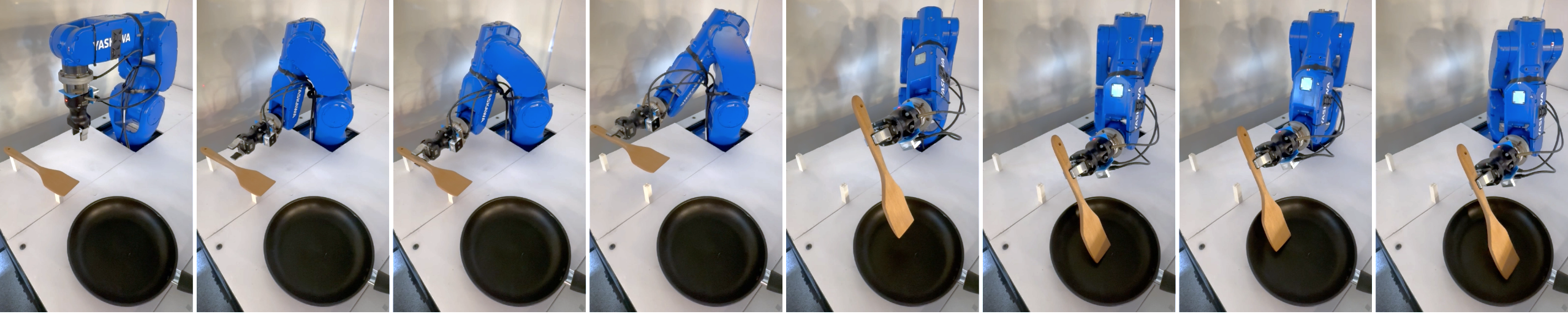}
    \includegraphics[width=\linewidth]{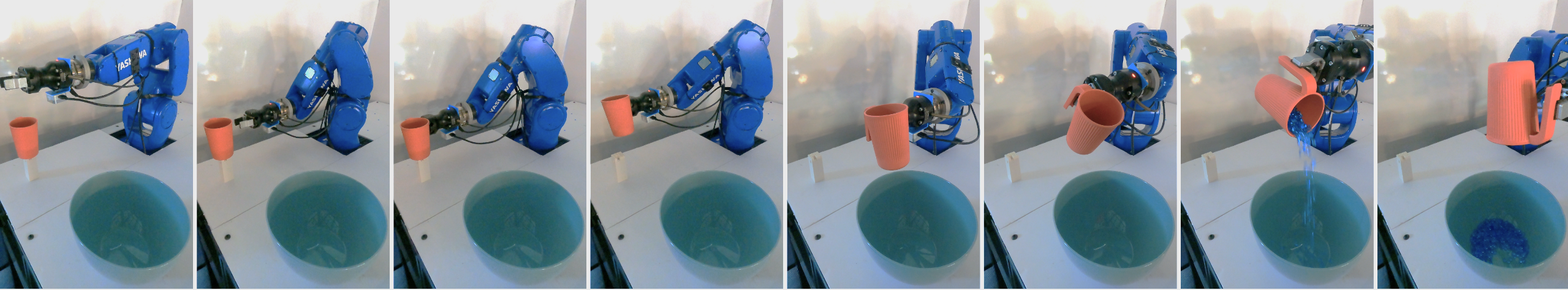}
    \includegraphics[width=\linewidth]{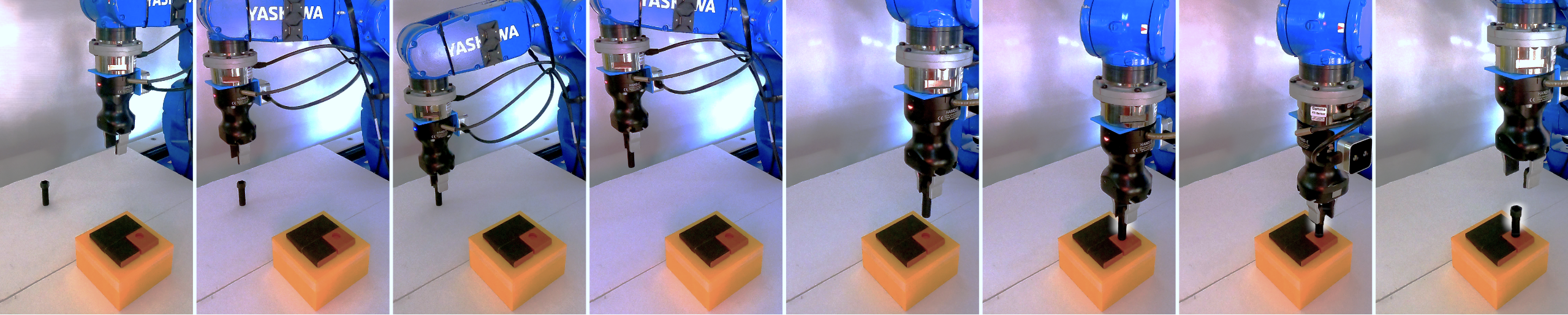}
    \includegraphics[width=\linewidth]{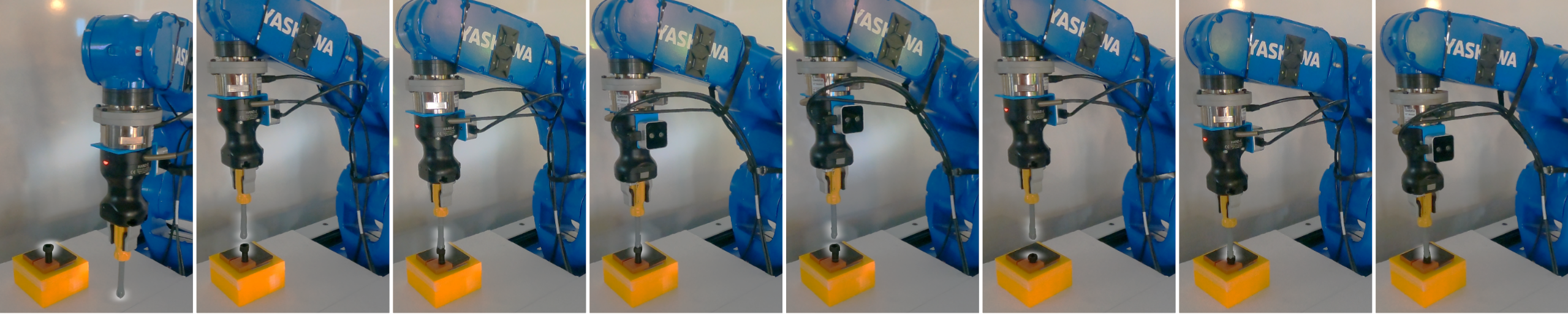}
    \caption{Pan scraping, pouring, screw insertion, screwing, respectively. Screws and screwdriver tips are highlighted for better visualization.}
    \label{fig:spatula_execute}
    \vspace{-0.5cm}
\end{figure*}

\section{Experiment Setup}
\label{sec:exp_setup}

We evaluate our method on three diverse manipulation tasks—pan scraping, pouring and whisking, and screwing—each representing a distinct application domain: cleaning, cooking, and industrial assembly. Despite the variability across these domains, all tasks follow a unified formulation: we define a small set of functionally meaningful keypoints for a set of reference objects, and derive task-axis controllers analytically using geometric relationships between keypoints and local surface geometry, allowing generalization without training.

In the pan scraping task, the robot grasps a spatula, aligns it with the given pan, and performs a simple scraping motion along the surface, following a direction projected from the pan handle to its center while maintaining contact force along the surface normal using force control. For the pouring and whisking task, the robot first grasps a mug, pours its contents into a bowl, and then grasps a whisk to stir the mixture inside. The screwing task involves grasping a screw, inserting it into a hole, and then grasping a screwdriver to drive it in while applying axial force along the screw’s center line. Each of these tasks involves multi-step interactions with multiple objects, yet the keypoint-to-controller mappings are consistent and modular across tools. For instance, the screwdriver and spatula both need to be positioned and aligned before a forceful interaction, enabling controller reuse. Similarly, the pan and bowl share surface-normal-based controllers centered at their base. Figure~\ref{fig:experiment_setup} presents visual examples of the tasks, object variants, keypoints, and GTAs defined for each object. 

To assess generalization, we collected an object data set covering a wide range of object appearances and poses. For each of the eight object classes, we selected three distinct object variants and placed them in three unique configurations varying in position, orientation, and robot viewpoint. Additionally, we collected three internet images per object class. This yielded 54 reference-target pairs per object class, totaling 432 pairs. All local RGB-D data were captured using an Intel RealSense D405 stereo camera mounted on the wrist of a Yaskawa GP4 6-DOF robotic arm equipped with a Robotiq Hand-E two-finger gripper. Examples of object variants, annotated keypoints, and the resulting task-axis controllers overlaid on point clouds are illustrated in Fig.~\ref{fig:experiment_setup}.

\section{Results}
\label{sec:result}
\subsection{Evaluating Keypoint Correspondences}

Before presenting quantitative results on grounding 3D task-axes, we first provide a qualitative evaluation of the predicted keypoint correspondences to illustrate the behavior and reliability of our keypoint matching. Figure~\ref{fig:controller_errors}A shows representative correspondence examples across a range of object categories from our dataset. In each image pair, the left image displays a reference example—either from our in-lab capture or collected from the internet—annotated with human-labeled keypoints. The right image shows the corresponding target object from our in-lab dataset, with keypoints predicted by our framework using SD-DINO features.

These examples highlight the model’s ability to accurately identify semantically aligned keypoints across significant variations in object configurations. While occasional misalignments arise in challenging settings, such as extreme viewpoint shifts, ambiguous geometry, or shadows (e.g., the screw example in the bottom right), the model generally demonstrates robust and consistent correspondence accuracy. 



\subsection{Evaluating Zero-Shot Task-Axis Groundings}

We evaluate the accuracy of our task-axis grounding pipeline by comparing the grounded task-axis controllers against ground-truth values from manual keypoint annotations. Figure~\ref{fig:controller_errors}B summarizes the position and rotation errors across different object classes and GTA types. Overall, the results show high accuracy across the board: positional controller errors are typically below 1~cm, and rotational errors remain under 3~degrees for most objects and tasks, demonstrating the effectiveness of our zero-shot controller grounding approach.

Notably, the apparent errors in some cases (e.g., the positional grasp controller on the spatula or screwdriver) reflect limitations of point-based evaluation rather than functional misalignment. The ground-truth grasp keypoint is a single manually labeled position, while in practice, a grasp may succeed anywhere along a functional region—such as the tool handle. Thus, small deviations in groundings do not necessarily translate to task failure. In fact, this tolerance to spatial variation suggests that our groundings are not only precise but also robust to annotation noise and ambiguity.

Figure~\ref{fig:controller_errors}B also provides insight into the impact of visual foundation model choice on controller accuracy. GTAs derived using only Stable Diffusion (light red and blue) show significantly higher variance, likely due to its limited semantic discrimination and reliance on low-level appearance features. DINOv2 (red and blue), which encodes functional and structural cues more effectively, performs better across all object classes. Our full model, SD-DINO (dark red and blue), consistently achieves the lowest error across all object classes, combining the semantic alignment of DINOv2 with the geometric refinement of Stable Diffusion. This confirms that the hybrid architecture of SD-DINO offers the strongest foundation for keypoint matching and, consequently, for accurate controller grounding. These results demonstrate that our framework enables reliable zero-shot extraction of object-centric task-axis controllers across diverse objects, highlighting the value of semantic correspondence and geometry-aware grounding in manipulation tasks.

\subsection{Transferring Semantically Similar Keypoints Across Distinct Objects}

This experiment demonstrates that our framework can transfer semantically meaningful keypoint configurations across different object categories. Both whisks and spatulas are annotated with the same keypoint settings. While they serve different purposes—mixing versus scraping—they share structural similarities: both are elongated handheld tools operated primarily through the tip. Leveraging this shared functional geometry, we directly reused the keypoint annotations from a reference whisk to predict corresponding keypoints on a set of novel spatulas. Figure~\ref{fig:transferring_controllers} (left) shows the annotated whisk, while the center image shows predicted keypoints on a sample spatula. From these correspondences, we grounded task-axis controllers for each spatula in our dataset. Figure~\ref{fig:transferring_controllers} (right) compares the task-axes errors when grounding spatula controllers from other spatulas versus the whisk. Although the whisk-based grounding yields slightly higher error, the performance remains within less than 1~cm difference for positions and 3 degrees of rotations—showcasing the robustness of our framework to cross-object generalization.

\subsection{Executing Controllers to Generalize Zero-Shot Skills}
We demonstrate the execution of multi-step manipulation skills using GTA controllers across three diverse tasks: pan scraping, pouring, and screw insertion and screwing. For each task, we defined a lifted skill composed of one or more GTA controllers based on the requirements of that specific manipulation and executed them in sequence after grounding on novel object instances.

Take the pan scraping task as an example. This skill involves several steps, including grasping the spatula, aligning it to the pan with a specific angle, and executing a scraping motion along the surface. Grasping using GTAs typically involves basic alignment of gripper with the object handles using \textsc{PosAlign} and \textsc{AxisAlign} controllers. Moving beyond grasping, the second step for aligning the tip of the spatula to the pan surface at a 45-degree angle from the surface normal and along the scraping direction, for example, we use the following three controllers in order of priority:
\begin{center}
\begin{minipage}{0.85\linewidth}
\raggedright
\textbf{\textsc{AxisAlign}}(spatula$_\text{tip\_dir}$, pan$_\text{surface\_dir}$, $[0,0,45]$) \\
\textbf{\textsc{AxisAlign}}(gripper$_\text{y}$, pan$_\text{scrape\_dir}$)\\
\textbf{\textsc{PosAlign}}(spatula$_\text{tip\_pos}$, pan$_\text{scrape\_pos}$)
\end{minipage}
\end{center}
These controllers ensure that the spatula tip is correctly oriented with respect to the pan surface while aligning with the functional scraping direction. The priority structure allows the axis alignment of the surface angle to take precedence, while the scraping direction is projected into the null space, enabling compliant motion across the pan’s surface. In the following step, the robot maintains contact force during the scraping motion using a \textsc{ForceAlign} controller along the spatula tip direction and moves along the scraping direction with \textsc{PosWaypoint}.

The pouring and screwing tasks were implemented in a similar fashion, using appropriate combinations of the controllers in our controller library. For the pouring task, the robot first grasps the mug using \textsc{PosAlign} and \textsc{AxisAlign} controllers, moves it with \textsc{PosAlign} to align with the bowl center, and then rotates the mug along its edge axis using an \textsc{AxisAlign} controller to pour its contents. The screw insertion task involves grasping and aligning a screw above the hole using  \textsc{PosAlign} and \textsc{AxisAlign} controllers, followed by pushing in using a \textsc{ForceAlign} controller along the insertion axis while rotating with \textsc{AxisAlign}. The screwing task uses similar controllers for alignment and adds force and rotational controllers to drive the screw in along its axis.

Figure~\ref{fig:spatula_execute} shows the real-world execution of all three tasks with sample objects. These executions demonstrate that our modular, zero-shot framework enables accurate and generalizable manipulation of novel objects using semantically grounded task-axis controllers. Notably, we were able to decompose all three multi-step tasks—despite their differences in tools, motions, and objectives—into low-level GTA controller skills instantiated from just four controller types. This compact yet expressive controller set highlights the versatility and reusability of our skill representation. In the future we plan to expand the set of controllers to support more precise manipulation control, e.g., controlled pouring.

\section{Conclusion}
\label{sec:conclusion}

We presented a zero-shot framework for transferring manipulation skills by grounding modular task-axis controllers using semantically matched keypoints. By leveraging visual foundation models like SD-DINO, our method enables accurate grounding of controller parameters across novel objects without requiring retraining or demonstrations. We demonstrated reliable generalization of complex, multi-step tasks such as scraping, pouring, and screwing, across varied object categories and configurations. Our approach offers a modular, interpretable path toward scalable skill reuse and generalization in open-world robotic manipulation.


\bibliographystyle{IEEEtran}
\bibliography{main}  


\end{document}